\def\year{2020}\relax
\documentclass[letterpaper]{article} 
\usepackage{aaai20}  
\usepackage{times}  
\usepackage{helvet} 
\usepackage{courier}  
\usepackage[hyphens]{url}  
\usepackage{graphicx} 
\def\UrlFont{\rm}  
\usepackage{graphicx}  
\newcommand{\citet}[1]{\citeauthor{#1}\shortcite{#1}}
\newcommand{\citep}{\cite}

\usepackage{amsmath}


\usepackage{multirow}    

\usepackage{amsfonts} 
\usepackage{subfigure} 

\author{\Large \textbf{Jie Yang, Zhiquan Qi\thanks{Corresponding author}, Yong Shi\thanks{Corresponding author}}\\
	University of Chinese Academy of Sciences\\
	Beijing 100190, China\\
	{\tt\small yangjie181@mails.ucas.ac.cn, qizhiquan@foxmail.com, yshi@ucas.ac.cn}
}

\title{Learning to Incorporate Structure Knowledge for Image Inpainting}

\begin{document}

\maketitle

\begin{abstract}

  This paper develops a multi-task learning framework that attempts to incorporate the image structure knowledge to assist image inpainting, which is not well explored in previous works. The primary idea is to train a shared generator to simultaneously complete the corrupted image and corresponding structures --- edge and gradient, thus implicitly encouraging the generator to exploit relevant structure knowledge while inpainting. In the meantime, we also introduce a structure embedding scheme to explicitly embed the learned structure features into the inpainting process, thus to provide possible preconditions for image completion. Specifically, a novel pyramid structure loss is proposed to supervise structure learning and embedding. Moreover, an attention mechanism is developed to further exploit the recurrent structures and patterns in the image to refine the generated structures and contents. Through multi-task learning, structure embedding besides with attention, our framework takes advantage of the structure knowledge and outperforms several state-of-the-art methods on benchmark datasets quantitatively and qualitatively.

\end{abstract}

\section{Introduction}

Image inpainting aims at filling corrupted or replacing unwanted regions of images with plausible and fine-detailed contents, which is widely applied in fields of restoring damaged photographs, retouching pictures, et al.

Existing inpainting approaches can be roughly divided into two groups: conventional and deep learning based approaches. Conventional inpainting approaches usually make use of low-level features (e.g. color and texture descriptors) hand-crafted from the incomplete input image and resort to priors (e.g. smoothness and image statistics) or auxiliary data (e.g. external image databases). They either propagate low-level features from surroundings to the missing regions following a diffusive process~\cite{diffusion1,diffusion3,diffusion4}~or fill holes by searching and fusing similar patches from the same image or external image databases~\cite{patch2,patch3,mrf1,mrf2}. Without a high-level understanding of the image contents and structures, conventional approaches usually struggle to generate semantically meaningful content, especially when a large portion of an image is missing or corrupted.
\begin{figure}
  \centering
    \includegraphics[width=1.\columnwidth]{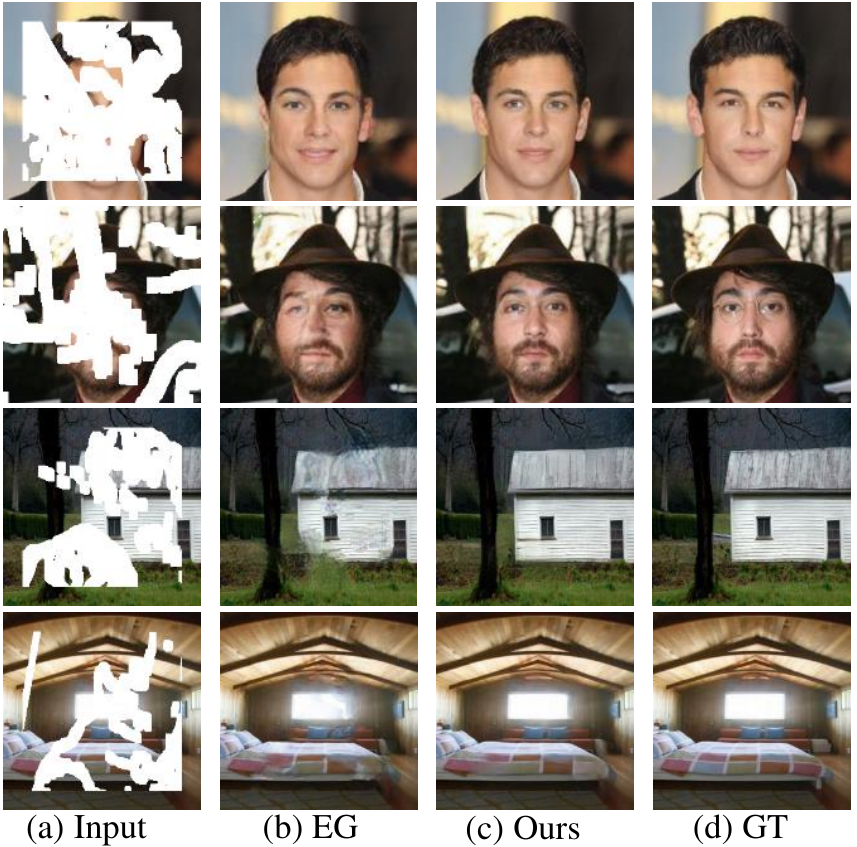}
    \caption{Our results compared with EG~\cite{EG} which exploits structure knowledge with a series-coupled architecture and the ground truth (GT). [Best viewed with zoom-in.]}
    \label{edge_ours}
\end{figure}

Deep learning-based approaches can understand the image content by automatically capturing the intrinsic hierarchical representations and generate high-level semantic features to synthesize the missing contents, which generally outperform the conventional methods in the inpainting task. Context Encoder proposed by~\citet{CE}~is the first attempt to exploit a deep convolution encoder-decoder trained with an adversarial strategy for image inpainting. The method produces semantic reasonable contents, but the results often lack fine-detailed textures and contain visible artifacts.
To achieve more pleasing results,~\citet{GLI,NPS,CA,Patch-Swap,Shift-Net}~and~\citet{NIPS}~respectively extend Context Encoder in different ways, such as in the aspects of architectures and learning strategies.



Recently,~\citet{EG}~propose to utilize explicit image structure knowledge for inpainting. They develop a two-stage model which comprises of an edge generator followed by an image generator. The edge generator is trained to hallucinate the possible edge sketches of the missing regions. Then the image generator makes the generated sketches as a structure prior or precondition to produce final results.~\citet{Contour}~propose a similar model but take a contour generator instead of an edge generator which is more applicable in the cases where the corrupted image contains salient objects. By introducing the structure information, both methods generate more visually plausible inpainting results.

The success of the above two-stage models suggests that structure knowledge such as edges and contours plays an important role to generate reasonable and detailed contents for image inpainting. It also indicates that, without advisable guidance of structure knowledge in the learning process, previous deep learning-based approaches may struggle to understand the plausible semantic structures of the corrupted images. However, the two-stage strategy may suffer several limitations: 1) it takes much more parameters since using two generators; 2) it is easy subjected to the adverse effects from unreasonable structure preconditions during the inference time due to using a series-coupled architecture; 3) without an explicit structure guidance as a loss function during the learning process, it may not sufficiently incorporate the structure information since they may be weakened or forgotten due to the sparsity of the structures and the depth of the network.

Based on these insights, we propose to use a multi-task framework to better incorporate structure knowledge for image inpainting. Instead of explicit modeling the structure preconditions, we utilize a shared generator to simultaneously generate the completed image and corresponding structures, thus supervising the generator to incorporate relevant structure knowledge for inpainting. This is reasonable because both tasks require a high-level understanding and share the same semantics of the image content. Besides,~\citet{EG}~and~\citet{Contour}~have demonstrated that structure priors are benefiting to image completion; the other way round, it is more likely to figure out the complete structures from a relatively intact image compared with a corrupted one.


In addition, to further incorporate the structure information, we introduce a structure embedding scheme which explicitly feeding the learned structure features into the inpainting process serving as preconditions for image completion. Moreover, an attention mechanism is developed to exploit the recurrent structures or patterns in the image to refine the generated structures and contents. Specifically, we also propose a novel pyramid structure loss to supervise the learning of the structure knowledge. We summarize the main contributions as follows:
\begin{itemize}
\item We propose a multi-task learning framework to incorporate the image structure knowledge to assist image inpainting.
\item We introduce a structure embedding scheme which can explicitly provide structure preconditions for image completion, and an attention mechanism to exploit the similar patterns in the image to refine the generated structures and contents.
\item We propose a novel pyramid structure loss specifically for structure learning and embedding. Extensive experiments have been conducted to evaluate the performance of our approach.
\end{itemize}

\section{Related Work}
Numerous image inpainting approaches have been proposed; here, we focus to review the representative deep learning-based methods.


Context Encoder proposed by~\citet{CE}~is one of the first deep learning-based methods for image inpainting, which takes an encoder-decoder architecture and trains with an adversarial learning strategy. It leverages convolutional encoder-decoder and Generative Adversarial Network~\cite{gan}, thus able to develop semantic features and synthesis visually pleasing contents even the missing regions are quite large. But the inpainting results often lack fine-detailed textures due to the information bottleneck layer of the encoder-decoder which may discard some features for image details. Besides, the approach tends to create artifacts around the border of the missing region due to the local consistency is not taken into consideration.


\citet{GLI}~address the information bottleneck defect by replacing the bottleneck layer with a series of dilated convolution layers and reducing the downsampling times. For local continuity, a local discriminator is designed to enforce the locally filled content is both visually plausible and consistent with the surroundings. Although the method can plausibly fill missing regions, it still takes Poisson blending~\cite{Poisson}~to tackle the color inconsistency between the completed region and its surroundings.~\citet{NPS}, in a different way, enhance Context Encoder by proposing a multi-scale neural patch synthesis approach. The approach first takes the output of the network as initialization and then leverages style transfer techniques~\cite{ST}~to propagate the high-frequency textures from the surroundings to the missing region by iteratively solving a multi-scale optimization problem. The approach works well for high-resolution semantic inpainting.

~\citet{CA}~propose a two-stage coarse-to-fine architecture to generate and refine the inpainting results, where the coarse network makes an initial estimation, and the refinement network takes the initialization to produce finer results. Besides, at the refinement stage, a novel module termed as Contextual Attention is designed to explicitly borrowing information from the surroundings of the missing regions.~\citet{Patch-Swap}~develop a similar coarse-to-fine method and introduce a Patch-Swap module which can heuristically propagate the textures from surroundings to the holes. The coarse-to-fine architecture does help to generate finer results; however, it builds upon the assumption that the coarse estimate at the first stage is reasonably accurate. Similar to the ideas of Context Attention and Patch-Swap,~\citet{Shift-Net}~develop a shift-connect module by which the features of the known background at the encoding phase are directly shifted to fill the missing areas at the decoding phase. Unlike using an explicit module to propagate information from the surroundings to missing regions,~\citet{NIPS}~introduce an implicit diversified Markov random fields (ID-MRF) loss which implicit constraints the network to propagates relevant information to the target inpainting areas. And to leverage features of both image-level and feature-level,~\citet{PEN-Net} propose a pyramid-context encoder network and an attention transfer mechanism which are able to progressively fill the missing regions from high-level to low-level feature map and ensure the semantic consistencies at the same time.

To generalize well in the inpainting tasks of irregular missing regions,~\citet{Partial}~propose partial convolutions. Unlike vanilla convolution, partial convolution only utilizes valid information to inference the missing contents through an automatic mask updating mechanism which is effective in cases of arbitrary missing regions.~\citet{CA2}~further generalize the partial convolution and propose a gated convolution with a learnable mask updating mechanism which achieve competitive or better inpainting qualities. Besides, the users are able to interact with the inpainting network with hand-drawn sketches to produce user-guided inpainting results.

Recently, several approaches explicitly introduce image structure prior (e.g. edges and contours) for inpainting which produce more impressive results.
~\citet{EG}~propose a model termed as EdgeConnect which consists of an edge generator followed by an image generator. The edge generator is utilized to estimate the possible edges of the missing region, which then as precondition information feed into the successive image completion process.~\citet{Contour}~develop a similar model which takes a contour generator instead of the edge generator. Since the approach predicts contours for salient objects, it is more applicable in the cases where the corrupted image contains salient objects.

\section{Method}

\begin{figure*}
  \centering
    \includegraphics[width=1.\linewidth]{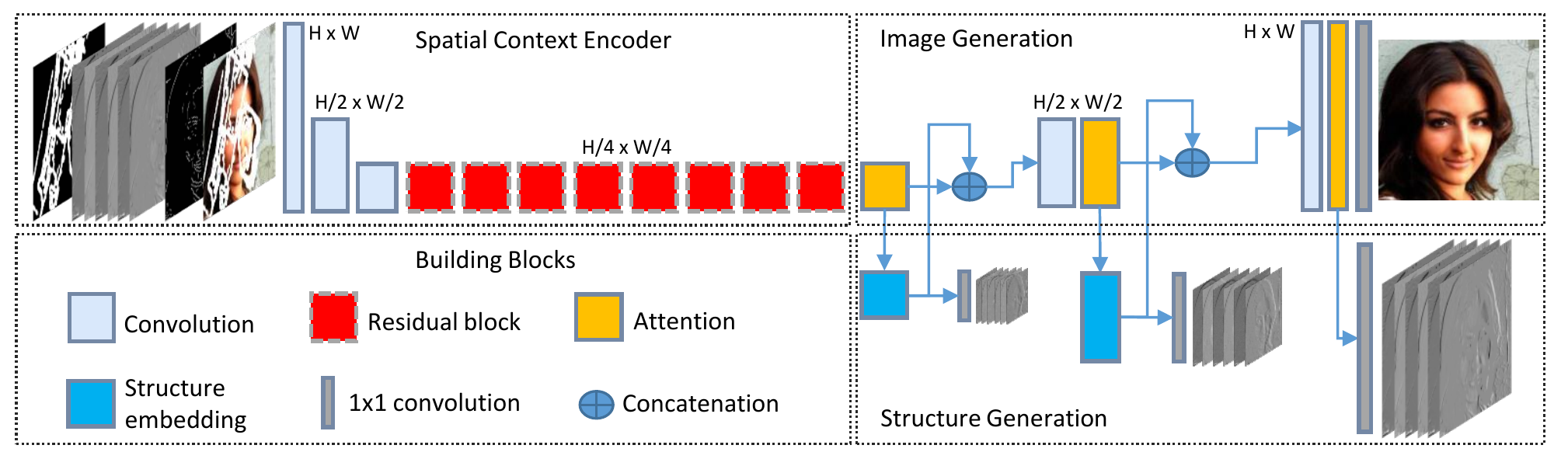}
    \caption{The overview of our multi-task framework. It leverages the structure knowledge with multi-tasking learning (simultaneous image and structure generation), structure embedding and attention mechanism. [Best viewed in color.]}
    \label{img-architecture}
\end{figure*}

Our multi-task framework is shown in Figure~\ref{img-architecture}. It estimates a shared generator for simultaneously generating the complete image and corresponding structures at different scales, where the structure generation works as an auxiliary task providing possible structure cues for the image completion task.

Here, we mainly use the edge structures to represent the image structure which describe the profiles of the contents of the image. Instead of directly figuring out the possible edges, we first predict the whole gradient map which inherently contains the edge information and then introduce an implicit regularization scheme in the proposed pyramid structure loss to learn the edge structures. Generating the gradient map is preferable in our multi-task setting. One the one hand, since the edge structure of an image is usually sparse and only conveys binary sketch information of the image, generating such edge structure shares little features with the task of image generation during the last several phases of the generation process, thus task-specific network layers for edge generation have to be designed. One the other hand, the gradient map itself not only conveys the possible edge information but also represents the texture information or high-frequency details which is important for detailed texture synthesis~\citet{grad1,grad2}.

\begin{figure}
  \centering
    \includegraphics[width=0.65\columnwidth]{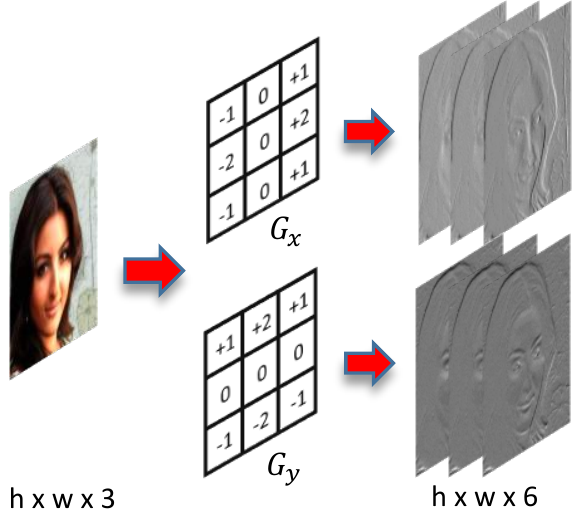}
    \caption{Gradient map ($h\times{w}\times{6}$) extracted from a RGB image with size of~($h\times{w}\times{3}$)~by Sobel filters~$G_x$~and~$G_y$.}
    \label{img-sobel}
\end{figure}

Formally, let's $\mathbf{I}$ be the ground truth image, $\mathbf{C}$ and $\mathbf{E}$ denote its gradient and edge map respectively. Here, we use Sobel filters shown in Figure~\ref{img-sobel} to extract the gradient map, and Canny detector to acquire the edge map.

The generator takes the masked image $\mathbf{\hat{I}=I\odot{(1-M)}}$ as the input, and corresponding gradient map $\mathbf{\hat{C}=C\odot{(1-M)}}$ and edge map $\mathbf{\hat{E}=C\odot{E}}$, in addition with the image mask $\mathbf{M}$ (with value 0 for known region 1 otherwise) as preconditions. Here, $\odot$ denotes the Hadamard product. The generator jointly generates the image content and estimates its gradient map at different scales:
\begin{equation}
(\mathbf{I}_{pred},\mathbf{C}_{pred}^{(s)})=G(\mathbf{\hat{I}},\mathbf{\hat{C}},\mathbf{\hat{E}},\mathbf{M})
\end{equation}
where $G$ represents our generator, $\mathbf{I}_{pred}$ the generated image, $\mathbf{C}_{pred}^{(s)}$ denotes the predicted gradient map at scale $s$. The final completed image and gradient map are $\mathbf{I}_{comp}=\mathbf{\hat{I}}+\mathbf{I}_{pred}\odot{\mathbf{M}}$ and $\mathbf{C}_{comp}^{(s)}=\mathbf{\hat{C}}^{(s)}+\mathbf{C}_{pred}^{(s)}\odot{\mathbf{M}^{(s)}}$, where $\mathbf{\hat{C}}^{(s)}$ is the incomplete gradient map at scale $s$. The number of scales is upon the specific architecture of the generator.


\subsection{Architecture}

We take the architecture proposed by~\citet{EG} as the backbone of our generator, which has achieved impressive results for image inpainting.
As Figure~\ref{img-architecture} shows, for image generation, the generator consists of a spatial context encoder which down-samples twice followed by eight residual blocks and a decoder which up-samples twice to generate images of the original size. For structure generation, the encoder is shared and the decoder is adapted to a multi-scale style to embed and output the structures of different scales. In addition, two modules are developed to make use of the structure information:





\paragraph{Structure Embedding Layer} We use the structure embedding layers to embed the structure features into the decoding phase at different scales serving as priors for image generation. It first separates from the image generation branch to learn the specific structural features and predict the possible structures, then merges the learned features back through a concatenation operation. This parallel/sibling-style scheme not only provides the structure priors for image generation but also avoids the adverse effects from improper preconditions since the decoder can learn to whether to exploit the structure priors or not. Specifically, we implement the layer with a standard residual block~\cite{res}.


\paragraph{Attention Layer} Our attention operation is inspired by the non-local mean mechanism which has been used for deionizing~\cite{non_local1} and super-resolution~\cite{non_local2}. It calculates the response at a position of the output feature map as a weighted sum of the features in the whole input feature map. And the weight or attention score is measured by the feature similarity. Through attention, similar features from surroundings can be transferred to the missing regions to refine the generated contents and structures (e.g. smoothing the artifacts and enhancing the details).






Given an input feature map, we first extract the feature patches and calculate the cosine similarity $s_{i,j}$ of each pair of the patches:
\begin{equation}
s_{i,j}=\langle\frac{p_i}{||p_i||_2}, \frac{p_j}{||p_j||_2}\rangle
\end{equation}
where $p_i$ and $p_j$ are the $i$-th and $j$-th patch of the input feature map $\mathbf{x}$ respectively. Then softmax operations are applied to compute the attention scores:
\begin{equation}
\hat{s}_{i,j}=\frac{e^{s_{i,j}}}{\sum_{j=1}^{m}e^{s_{i,j}}}
\end{equation}

Supposing a total of $m$ patches are extracted, the response of a position $o_{i}$ in the output feature map is calculated as the weighted sum of the patch features:
\begin{equation}
o_{i}=\sum_{j=1}^{m}{\hat{s}_{i,j}p_{j}}
\end{equation}

\begin{figure}
  \centering
    \includegraphics[width=1.\columnwidth]{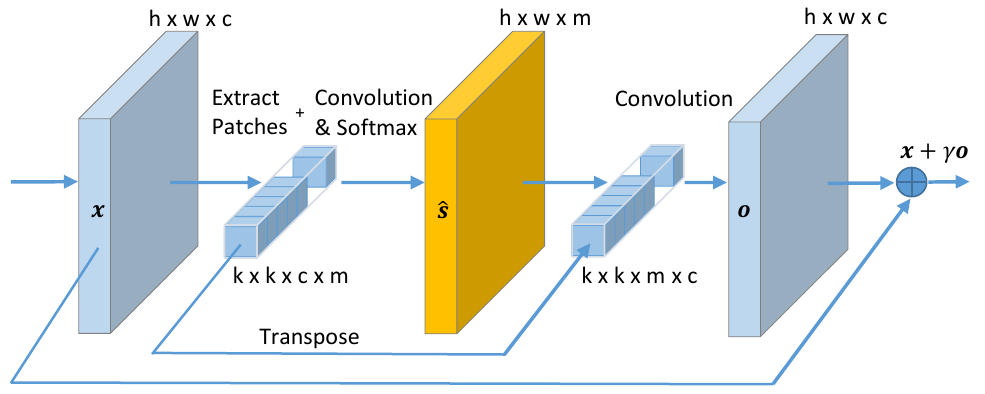}
    \caption{The proposed attention layer. It extracts $m$ feature patches as convolution filters with shape  ($k\times{k}\times{c}$) from the input feature map $\mathbf{x}$ ($h\times{w}\times{c}$) and computes the attention score maps $\mathbf{\hat{s}}$ through convolutions between the filters and the input followed by softmax operations, then convolves the scores back to reconstruct the feature map $\mathbf{o}$, finally adds it back to the input feature map with a scale parameter $\gamma$.}
    \label{attention}
\end{figure}

In particular, as shown in Figure~\ref{attention}, we formulate all the operations into convolution forms, and make it a residual block which thus can be seamlessly embedded into our architecture:
\begin{equation}
\mathbf{y}=\mathbf{x}+{\gamma}\mathbf{o}
\end{equation}
where $\mathbf{y}$ is the residual output, $\gamma$ is a learnable scale parameter.

\subsection{Loss Functions}
Our generator is expected to achieve two goals --- figuring out the structure cues and completing the corrupted image. We introduce a pyramid structure loss to capture the structure knowledge and a hybrid image loss to supervise image inpainting.

\paragraph{Pyramid Structure Loss} We propose a pyramid structure loss to guide the structure generation and embedding, thus incorporating the structure information into the generation process.
Specifically, it consists of two terms at a specific scale $s$. One is the $L^1$ distance between the predicted gradient map and corresponding ground truth, the other is a regularization term for learning the edge structure:

\begin{equation}
\mathcal{L}_{structure}=\sum_{s}^{n_s}[||\mathbf{C}_{pred}^{(s)}-\mathbf{C}^{(s)}||_{1}+\beta\mathcal{L}_{edge}^{(s)}]
\end{equation}
where $\mathcal{L}_{edge}^{(s)}$ denotes the regularization term, $\beta$ corresponding coefficient and $n_s$ the number of total scales. To implement the regularization on the edge structure,
we first use a Gaussian filter $g$ to convolve the binary ground truth edge map $\mathbf{E}^{(s)}$ to create a weighted edge mask as:
\begin{equation}
\mathbf{M}_{E}^{(s)}=g*\mathbf{E^{(s)}}
\end{equation}
Then, we computes the edge regularization loss as:
\begin{equation}
\mathcal{L}_{edge}^{(s)}=||\mathbf{C}_{pred}^{(s)}-\mathbf{C}^{(s)}||_{1}\odot{\mathbf{M}_{E}^{(s)}}
\end{equation}
where the weighted edge mask is used to extract the edge information from the gradient map. Using such an edge mask not only considers the positions of the binary edges but also exert constraints on their nearby locations, thus to highlight and intensify the edge structure. In our implementation, a Gaussian filter with size $10\times{10}$ and standard deviation $1$ is used.


\paragraph{Hybrid Image Loss} We take a similar hybrid loss as in~\cite{EG} for image completion, which consists of a pixel-wise reconstruction loss, a perception loss, a style loss and an adversarial loss which are detailed as follows.

The reconstruction loss is measured by the $L^1$ distance between the generated image~$\mathbf{I}_{pred}$~and corresponding ground truth at pixel level:
\begin{equation}
\mathcal{L}_{rec}=||\mathbf{I}_{pred}-\mathbf{I}||_{1}
\end{equation}

The perceptual loss computes the $L^1$ distance between~$\mathbf{I}_{pred}$~and its ground truth in the feature spaces after feeding to the pre-trained VGG-19 network~\cite{vgg19}~on ImageNet dataset~\cite{imagenet}.
\begin{equation}
\mathcal{L}_{perc}=\sum_{i}||\phi_{i}(\mathbf{I}_{pred})-\phi_{i}(\mathbf{I})||_{1}
\end{equation}
where~$\phi_i$~is the feature map of the~$i$'th selected layer from VGG-19. Here, layers~$relu1\_1$,~$relu2\_1$,~$relu3\_1$,~$relu4\_1$~and~$relu5\_1$ are used.

Style loss also compares the $L^1$ distance between images in feature spaces, but first computing corresponding Gram matrix~\cite{ST}~of each selected feature map:
\begin{equation}
\mathcal{L}_{style}=\sum_{i}||G_{\phi_{i}}(\mathbf{I}_{pred})-G_{\phi_{i}}(\mathbf{I})||_{1}
\end{equation}
where~$G_{\phi_{i}}$~is a $C_i\times{C_i}$~Gram matrix constructed from feature maps~$\phi_{i}$~of size~$H_i\times{W_i}\times{C_i}$.

In our framework, an adversarial training strategy is also used which almost has been a standard practice in image generation tasks. We take PatchGAN~\cite{img_img}~as our discriminator~$D$~and denote its adversarial loss as:
\begin{equation}
\mathcal{L}_{D}=\mathbb{E}_{\mathbf{I}}[logD(\mathbf{I})]\\+\mathbb{E}_{\mathbf{I}_{comp}}log[1-D(\mathbf{I}_{comp})]
\end{equation}
and the adversarial loss for our generator as:
\begin{equation}
\mathcal{L}_{G}=\mathbb{E}_{\mathbf{I}_{comp}}log[1-D(\mathbf{I}_{comp})]
\end{equation}

Then, the hybrid image loss $\mathcal{L}_{image}$ is defined as:
\begin{equation}
\mathcal{L}_{image}=\mathcal{L}_{rec}+\lambda_{1}\mathcal{L}_{prec}+\lambda_{2}\mathcal{L}_{style}+\lambda_{3}\mathcal{L}_{G}
\end{equation}
where~$\lambda_{1}$,~$\lambda_{2}$~and~$\lambda_{3}$~are hyperparameters which balance the contributions of different loss terms.

Finally, the generator is optimized by minimizing the pyramid structure loss and the hybrid image loss:
\begin{equation}
\mathcal{L}=\mathcal{L}_{image}+\alpha\mathcal{L}_{structure}
\end{equation}
where~$\alpha$~is a predefined weight to balance the two learning tasks. For our experiments, we choose hyperparameters of the hybrid image loss as in~\cite{EG}, and $\alpha=0.1$, $\beta=100$.

\section{Experiments}
In this section, we present our experimental comparisons with several state-of-the-art image inpainting approaches and ablation studies of the effectiveness of our multi-task framework. More results can reference our supplementary material.

\subsection{Experimental Settings}

\paragraph{Datasets and Baslines}
We evaluate our approach on three datasets of CelebA~\cite{celeba}, Places2~\cite{places2} and Facade~\cite{facade}, and compare the results with the following state-of-the-art methods both qualitatively and quantitatively:

\begin{itemize}
\item[-] GL: proposed by~\citet{GLI}, which uses two discriminators to ensure global and local consistency of the generated image.
\item[-] CA: proposed by~\citet{CA}, which leverages a coarse-to-fine architecture with a contextual attention layer to produce and refine the inpainting results.
\item[-] PEN-Net: proposed by~\citet{PEN-Net}, which adopts a pyramid context encoder to fill missing regions with features of both image-level and feature-level.
\item[-] EG: proposed by by~\citet{EG}, which leverages the edge structure preconditions for inpainting with a series-coupled architecture.
\end{itemize}

We utilize the available pre-trained models of the baseline approaches and reimplement PEN-Net~\cite{PEN-Net}~as there is no publicly available code yet.



\paragraph{Implementation Details}
For experiments, we resize images to $256\times{256}$ and use both regular and irregular image masks for training and testing. For fair comparisons, we use regular masks (with a size of $128\times{128}$) following the common experimental settings of baselines and irregular masks as in baseline~\cite{EG}. We generate gradient maps with Sobel filters and edge maps with Canny detectors as in~\cite{EG}. To compute the pyramid structure loss, we scale these maps into corresponding resolutions with nearest-neighbor interpolation. We implement our model in TensorFlow using a single NVIDIA GeForce GTX 1080 Ti and the code will be publicly available.

\begin{figure}
  \centering
    \includegraphics[width=0.85\columnwidth]{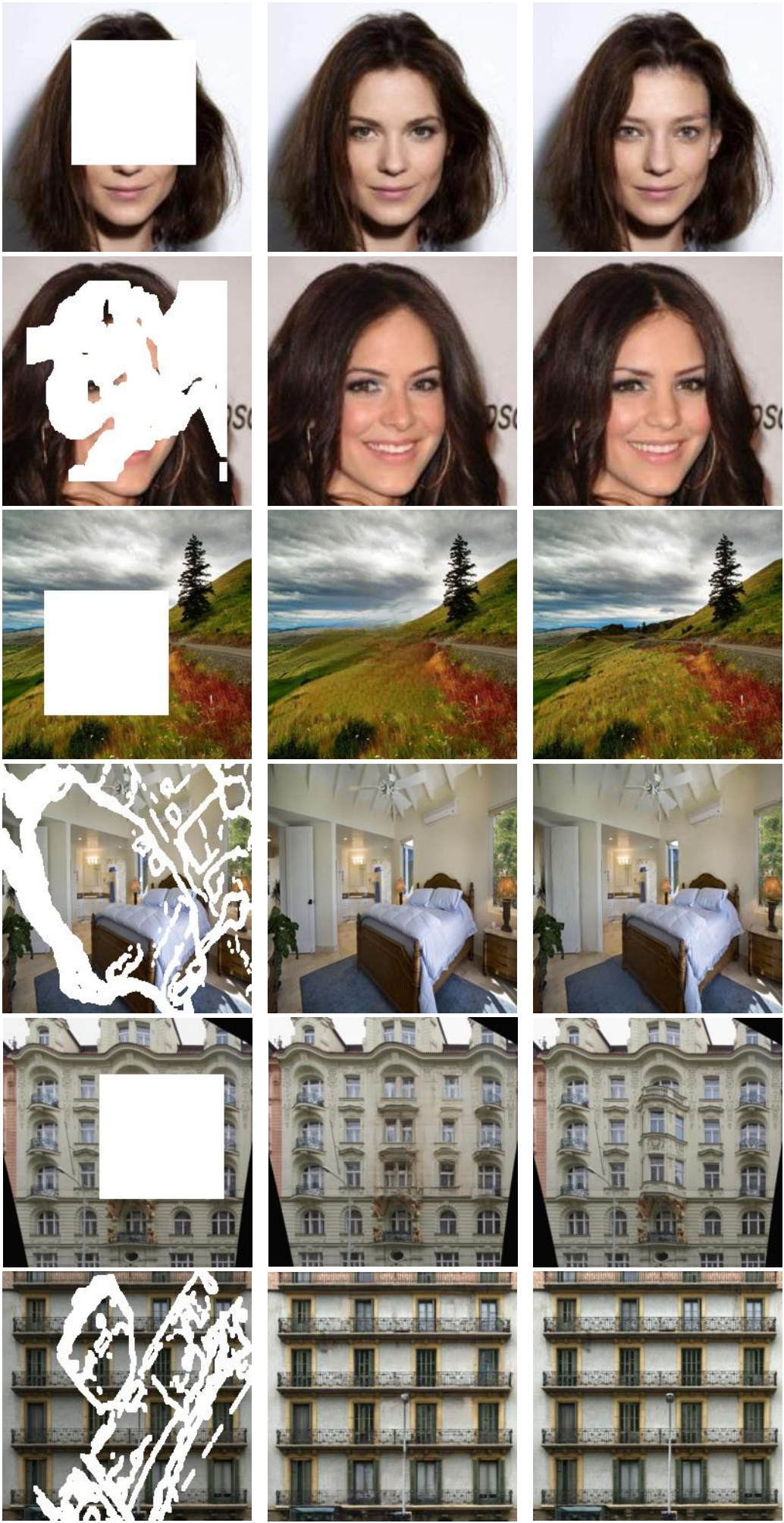}
    \caption{Example inpainting results on CelebA, Places2 and Facade. From left to right: Input, Ours and Ground truth. [Best viewed with zoom-in.]}
    \label{quality_results}
\end{figure}
\subsection{Qualitative Evaluation}
\begin{figure}[htbp]
\centering
    \includegraphics[width=1.\linewidth]{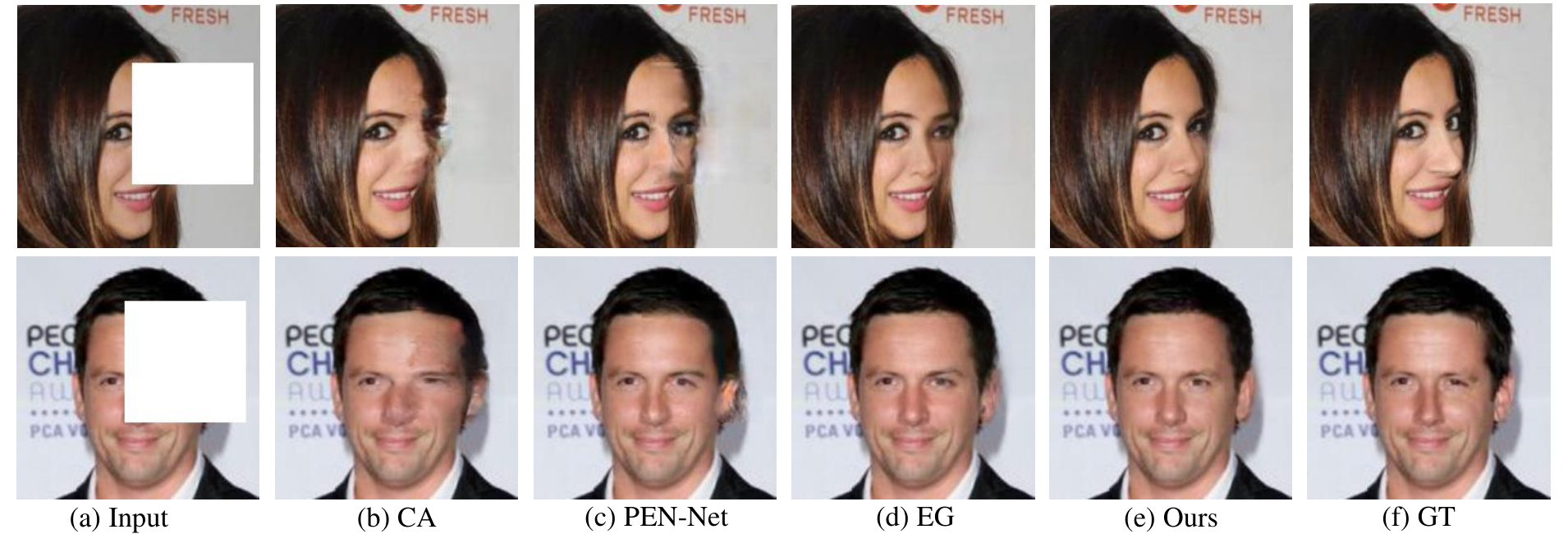}
\caption{Qualitative comparisons with baselines and the ground truth (GT) on CelebA. [Best viewed with zoom-in.]}
\label{celeba}
\end{figure}

\begin{figure}[htbp]
\centering
    \includegraphics[width=1.\linewidth]{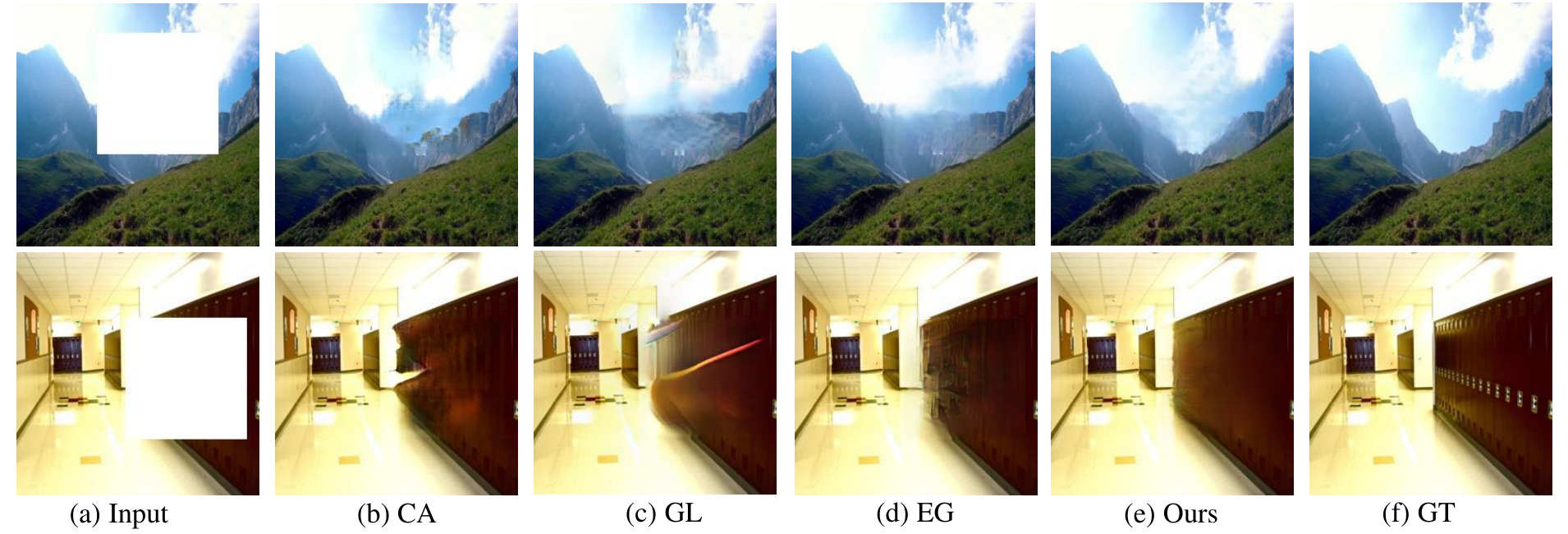}
\caption{Qualitative comparisons with baselines and the ground truth (GT) on Places2. [Best viewed with zoom-in.]}
\label{places2}
\end{figure}

As shown in Figure~\ref{quality_results}, our approach is able to generate visually realistic images with sharp edges and fine-detailed textures in both regular and irregular mask settings.
Besides, testing on regularly masked images as shown in Figure~\ref{celeba}~and Figure~\ref{places2}, ours compared with baselines shows obvious visual enhancement on pleasing image structures, such as sharp facial contours, crisp eyes and ears, and reasonable object boundaries. And comparing with the approaches CA, GL and PEN-Net where few image structure information is explicitly considered, ours and EG which incorporate the edge structure knowledge are more likely to generate plausible image contents.
Moreover, as shown in Figure~\ref{edge_ours}, Figure~\ref{celeba} and Figure~\ref{places2}, comparing against EG which using a serial-coupled architecture to exploiting structure knowledge, our multi-task architecture exhibits superior performance with more visually plausible structures and detailed contents.



\subsection{Quantitative Evaluation}
\paragraph{Numerical Metrics}
We take $L_{1}$ loss, peak signal-to-noise ratio (PSNR), structural similarity index (SSIM)~\cite{ssim}, universal quality index (UQI)~\cite{uqi}, visual information fidelity (VIF)~\cite{vif} and Frechet Inception Distance (FID)~\cite{fid} as our evaluation metrics.
Specifically, we utilize $l_{1}$ loss and PSNR to measure the similarity between two images at the pixel level, SSIM and UQI to assess the distortions of the generated image content relative to the ground truth, and VIF and FID to evaluate the overall visual quality, among which VIF correlates well with human perceptions and FID has been a commonly used metric for image generation. In addition, the metrics will be calculated over ten thousand random images in the test sets.



As shown in Table~\ref{tab_quality}, our approach achieves superior performance against all the baselines on datasets CelebA and Places2.
The results can be explained by the baseline approaches either ignore the structure knowledge of the image or not well make use of it. Besides, under the scenario with irregular masks, although models such as CA, GL, and PEN-Net can deal with irregular holes (like filling irregular holes with multiple regular patches), they usually show inferior performance since not particularly trained on irregular masks.






\begin{table}[h]
\centering
\caption{Quantitative comparisons with baselines.~$\P$~Lower is better.~$\dag$~Higher is better. [Best viewed with zoom-in.]}
\resizebox{.98\columnwidth}!{
\begin{tabular}{c c c c c c c c c}
\hline
\hline
Datasets                 & Masks                      & Models &$l_{1}\%\P$ &PSNR$\dag$ &SSIM$\dag$ &UQI$\dag$ &VIF$\dag$ &FID$\P$ \\ \hline
\multirow{8}{*}{Places2} & \multirow{4}{*}{irregular} & CA     &5.62        &21.95      &0.732      &0.939     &0.728     &27.81     \\ 
                         &                            & GL     &6.20        &22.40      &0.769      &0.939     &0.689     &19.03     \\ 
                         &                            & EG     &3.33        &24.97      &0.848      &0.967     &0.735     &13.74     \\ 
                         &                            & Ours   &\textbf{0.69}  &\textbf{27.07}  &\textbf{0.887}  &\textbf{0.975} &\textbf{0.787} &\textbf{4.883}     \\ \cline{2-9}

                         & \multirow{4}{*}{regular}   & CA     &4.44     &20.60      &0.773      &0.951     &0.732     &7.555     \\ 
                         &                            & GL     &4.91     &21.08      &0.777      &0.950     &0.697     &7.848     \\ 
                         &                            & EG     &3.90     &21.63      &0.786      &0.959     &0.709     &7.536     \\ 
                         &                            & Ours   &\textbf{3.52}  &\textbf{22.46}  &\textbf{0.813}  &\textbf{0.964} &\textbf{0.732} &\textbf{7.423}     \\ \hline \hline
\multirow{8}{*}{CelebA}  & \multirow{4}{*}{irregular} & CA     &4.87     &23.27      &0.790      &0.934     &0.805     &23.13     \\ 
                         &                            & PEN-Net&2.94     &28.02      &0.875      &0.972     &0.811     &10.42     \\ 
                         &                            & EG     &1.81     &30.44      &0.941      &0.979     &0.856     &2.443     \\ 
                         &                            & Ours   &\textbf{1.47}  &\textbf{33.19}  &\textbf{0.960}  &\textbf{0.985}  &\textbf{0.893} &\textbf{1.227}     \\ \cline{2-9}

                         & \multirow{4}{*}{regular}  & CA     &3.03     &23.51      &0.864      &0.962      &0.794     &4.033     \\ 
                         &                           & PEN-Net&2.54     &25.41      &0.905      &0.971      &0.802     &3.482     \\ 
                         &                           & EG     &2.39     &25.29      &0.901      &0.975      &0.809     &2.421     \\ 
                         &                           & Ours   &\textbf{2.08}   &\textbf{26.82}  &\textbf{0.927}  &\textbf{0.979}  &\textbf{0.842} &\textbf{1.654}     \\ \hline \hline
\end{tabular}}
\label{tab_quality}
\end{table}

\subsection{Ablation Study}

We analyze how the proposed components of our framework contribute to the final performance of image inpainting.
We take the image generator in~\cite{EG}~as the baseline, then gradually adding our multi-task learning strategy (MT), structure embedding (SE) and attention mechanism(AT) until establishing the whole model we proposed. Correspondingly, we evaluate the model with the  gradually added components quantitatively and qualitatively over one thousand random images in the test sets with regular masks.

\begin{figure}
\centering
    \includegraphics[width=1.\linewidth]{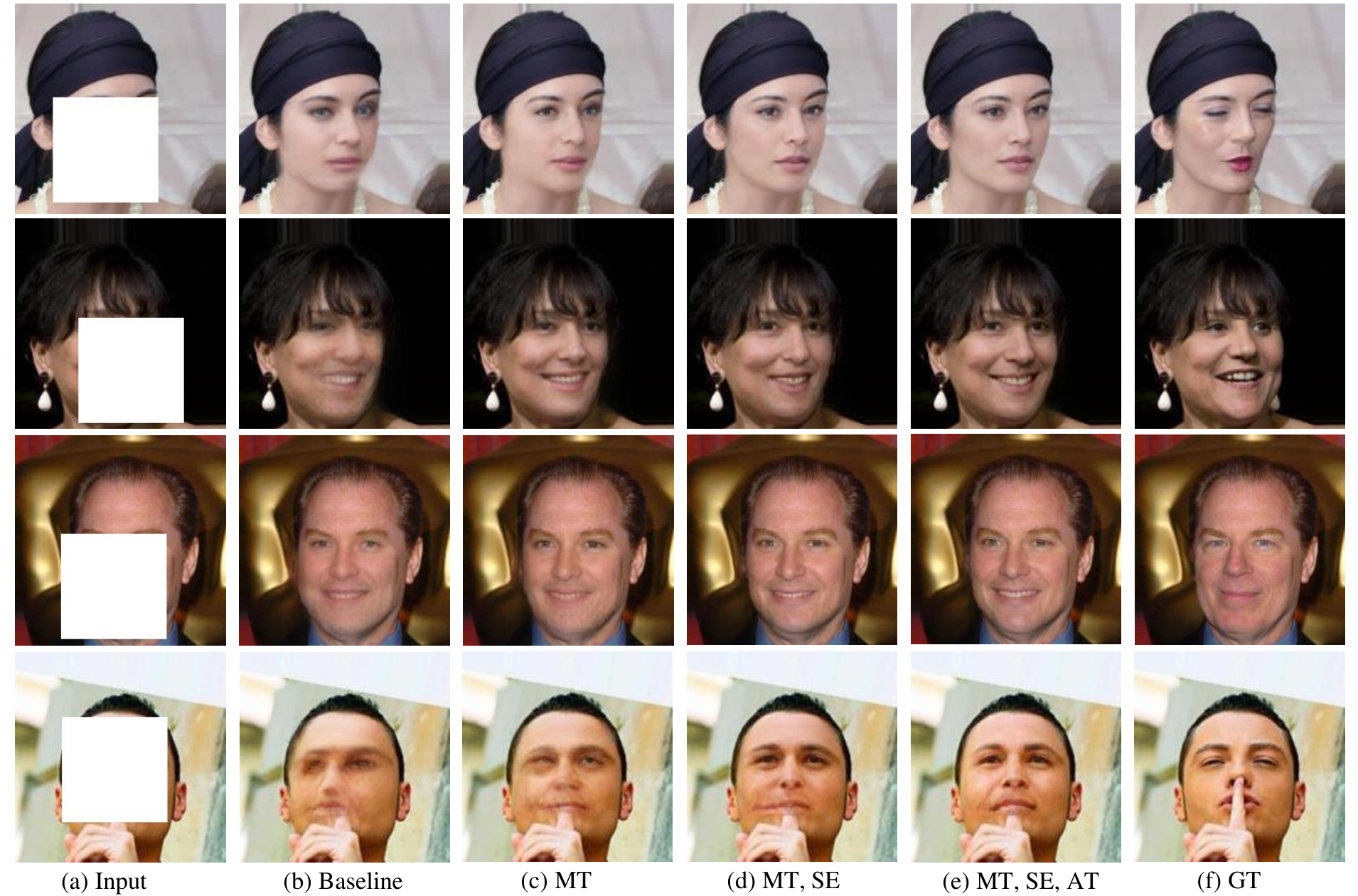}
\caption{Qualitative results of the ablation study. [Best viewed with zoom-in.]}
\label{ablation}
\end{figure}

As Table 2 shows, the performances of our model on the metrics are gradually improved or retained compared with the baseline as progressively integrating each component. Specifically, metric VIF and FID are enhanced by a large margin, which indicates the visual quality of the complete images are improved substantially.
As qualitative comparisons are shown in Figure~\ref{ablation},  when taking a shared generator to simultaneously complete the image and corresponding structures instead of the only image completion task as in baseline, ours generates more pleasing image structures (e.g. sharp facial and month contours), which suggests the proposed multi-task strategy shows great potentials for incorporating the structure knowledge into the inpainting process.
Besides, with the explicit embedding of the structure features, the inpainting results are further enhanced (e.g. more sharp contours and textures). Moreover, with the attention mechanism embedded, the results are finally polished by the similar structures and patterns in the images.



\begin{table}[h]
\centering
\caption{Quantitative results of the ablation study.~$\P$~Lower is better.~$\dag$~Higher is better. [Best viewed with zoom-in.]}
\resizebox{.98\columnwidth}!{
\begin{tabular}{lllllll}
\hline
\hline
Model configurations &$l_{1}\%\P$ &PSNR$\dag$ &SSIM$\dag$     &UQI$\dag$ &VIF$\dag$ &FID$\P$ \\ \hline
Baseline             &4.03        &26.50      &0.908          &0.977     &0.823     &15.25     \\ 
MT                   &3.83        &26.94      &\textbf{0.912} &0.978     &0.834     &13.66     \\ 
MT, SE               &3.84        &26.91      &0.903          &0.978     &0.844     &12.29     \\ 
MT, SE, AT           &\textbf{3.78} &\textbf{27.01} &0.911 &\textbf{0.979} &\textbf{0.848} &\textbf{11.98}  \\ \hline \hline
\end{tabular}}
\label{tab_ablation}
\end{table}

\section{Conclusion}

We have primarily presented a framework for incorporating image structure knowledge for image inpainting.  We propose to utilize the multi-task learning strategy, explicit structure embedding besides with an attention mechanism to make use of the image structure knowledge for inpainting. The experiments results demonstrate that the proposed approach shows superior performance compared with several state-of-the-art inpainting methods which either ignore or not well exploit the structure knowledge. Besides,  we have verified each proposed component for incorporating structure knowledge by ablation studies. In future work, we plan to investigate adapting the proposed multi-task framework to other specific inpainting architectures to leverage the structure knowledge.



\section*{Acknowledgements}
This work is supported by grants from: National Natural Science Foundation of China (No.71932008, 91546201, and 71331005).
\fontsize{9.0pt}{10.0}
\bibliography{inpainting}
\bibliographystyle{aaai}

\end{document}